\documentclass[runningheads]{llncs}
\usepackage[T1]{fontenc}
\usepackage{graphicx}
\usepackage{amsmath,bm}

\usepackage[hyphens]{url}
\usepackage{array}
\usepackage{diagbox}
\usepackage{adjustbox}
\usepackage{booktabs}
\begin{document}
\title{A Survey of Foundation Models for Environmental Science}
%
%

\author{Runlong Yu\textsuperscript{1}, Shengyu Chen\textsuperscript{1}\thanks{Runlong Yu and Shengyu Chen contributed equally.}, Yiqun Xie\textsuperscript{2}, \and Xiaowei Jia\textsuperscript{1}}

\authorrunning{R. Yu et al.}

\institute{University of Pittsburgh, Pittsburgh, USA 
    \and University of Maryland, College Park, USA \\
    \email{\{ruy59,shc160,xiaowei\}@pitt.edu, xie@umd.edu}
}

%
%
\maketitle              
\begin{abstract}
Modeling environmental ecosystems is essential for effective resource management, sustainable development, and understanding complex ecological processes. However, traditional methods frequently struggle with the inherent complexity, interconnectedness, and limited data of such systems. Foundation models, with their large-scale pre-training and universal representations, offer transformative opportunities by integrating diverse data sources, capturing spatiotemporal dependencies, and adapting to a broad range of tasks. This survey presents a comprehensive overview of foundation model applications in environmental science, highlighting advancements in forward prediction, data generation, data assimilation, downscaling, model ensembling, and decision-making across domains. We also detail the development process of these models, covering data collection, architecture design, training, tuning, and evaluation. By showcasing these emerging methods, we aim to foster interdisciplinary collaboration and advance the integration of cutting-edge machine learning for sustainable solutions in environmental science.

\end{abstract}
\section{Introduction}

Healthy environmental ecosystems are fundamental to human survival and well-being, providing essential resources such as clean air, water, food, and energy, which sustain life and fuel economic growth.  Modeling these systems is essential for understanding underlying processes and generating reliable predictions to guide resource management.  However, this is challenging because environmental systems are inherently complex, involving numerous interacting processes, and are often poorly observed due to the high costs of data collection. Traditionally, process-based physical models have been used to represent ecosystems across various environmental domains, including climate science, hydrology, agriculture, forestry, and geology~\cite{moorcroft2001method,hipsey2019general}. These methods rely on mathematical or physical equations to describe relevant processes. Given the complexity and limited knowledge in modeling certain processes, they often require approximations or parameterizations. Additionally, both calibration and inference for these models can be time-consuming and often demand significant domain expertise.

Advancements in data collection and processing in environmental science have sparked a growing interest in leveraging artificial intelligence (AI) and machine learning (ML)  to better model environmental ecosystems~\cite{bergen2019machine,karpatne2018machine}. These data-driven methods are particularly promising in scenarios where certain processes are not fully understood or require substantial computational resources. However, traditional ML models are typically designed for specific tasks, limiting their ability to capture the interconnectedness of various environmental processes. Integrating different ML models is challenging due to inconsistencies in model architectures, data inputs, and the scales of spatiotemporal processes involved. For instance, predicting water quality variables (e.g., water temperature and nutrient concentration) and water quantity variables (e.g., streamflow) is often handled by separate models, even though these variables are influenced by common processes. This siloed approach hinders understanding the relationships between tasks and limits effective information sharing across models. It may also introduce model bias due to 
the ignorance of related processes influencing the target variable. Additionally, clients (e.g, stakeholders and resource managers), who may lack expertise in AI and ML, often struggle to interpret and integrate results from multiple models to make informed decisions.

Recent ML studies have extensively explored different transfer learning and meta-learning strategies~\cite{zhuang2020comprehensive,hospedales2021meta}. These works have demonstrated the potential of many ML models to be transferred from data-sufficient tasks to target data-sparse tasks.  
Building on the concept of transfer learning, there is a rising trend toward developing foundation models \cite{bommasani2021opportunities}.  These models are pre-trained on a diverse set of tasks using either supervised or unsupervised methods to learn universal feature representations, enabling them to be fine-tuned for new tasks. Such power offers new opportunities for building data-driven models to address a broad range of environmental problems.  In particular,  many foundation models are able to harness large data from different sources and extract complex data patterns. They also offer flexibility in configuring input and output structures. For example, many large language models (LLMs) can accept user-specified inputs and generate different variables through proper prompt engineering. This is particularly useful in modeling environmental systems where we need to perform multiple related modeling tasks, but only a subset of them is observed in each data sample~\cite{luo2023free}. 
Additionally, foundation models can better adapt to new environments because they have been pre-trained on massive data~\cite{yu2025physics}.  

Recognizing the transformative potential of foundation models, this survey comprehensively reviews their applications in environmental science. To ensure thorough coverage, we draw on research from major academic databases, leading journals (e.g., \textit{Nature}, \textit{Water Resources Research}, \textit{Remote Sensing}, \textit{TGRS}, \textit{TPAMI}, \textit{TKDE}), top conferences (e.g., \textit{AAAI}, \textit{IJCAI}, \textit{NeurIPS}, \textit{ICML}, \textit{ICCV}, \textit{CVPR}, \textit{ACL}, \textit{KDD}, \textit{ICDM}, \textit{SDM}, \textit{WSDM}, \textit{PAKDD}), specialized workshops (e.g., \textit{KGML}), and preprints. We also review government reports from \textit{NASA}, \textit{USGS}, and \textit{NOAA} to capture the latest advances in large-scale environmental~AI.


The aim of this survey is to bring attention to the exciting advancements in the application of foundation models within environmental science, highlighting the opportunities for further research and development in this promising field. We hope that this survey will be valuable to both the machine learning community, by showcasing how foundation models are being developed to tackle complex environmental challenges, and to environmental scientists seeking to explore these cutting-edge models in their own work. 


We organize the paper as follows: Section~\ref{sec:background} provides an overview of the evolution from process-based models to foundation models, highlighting the paradigm shift in environmental modeling. Section~\ref{sec:objective} reviews the key objectives guiding the application of foundation models in environmental science. Section~\ref{sec:method} explores innovative methods and architectures for building these models. Section~\ref{sec:discussion} discusses future research opportunities, and the paper concludes with insights into the transformative potential of foundation models in environmental science. 

\section{Historical and Conceptual Overview}
\label{sec:background}

\subsection{Environmental Modeling}

The exponential growth of diverse environmental data has created unprecedented opportunities to advance our understanding of ecosystems. Traditional approaches to environmental computing have evolved through several stages~\cite{zeyang2022environmental}, each reflecting a distinct paradigm for addressing environmental challenges: 

\begin{itemize}
	\item \textbf{Process-based models (1.0):} These models are grounded in fundamental scientific principles, including physics, chemistry, and biology, to analyze the underlying mechanisms of environmental phenomena. Often referred to as process-based or theory-driven models, they rely on mathematical formulations, such as differential equations, to simulate key processes. Although these models are interpretable, they often struggle with complex systems characterized by high variability and limited observational data.
	
	\item \textbf{Data-driven models (2.0):} 
	With the rise of big data and AI, data-driven approaches emerged as a powerful alternative to traditional process-based models. These models prioritize pattern recognition, system characterization, and outcome prediction using large datasets. Aligning with the ``Fourth Paradigm'' of science, which emphasizes data-intensive methodologies, they excel at handling high-dimensional, complex problems. However, their ``black-box'' nature often limits interpretability and also requires large training data.
	
	\item \textbf{Hybrid physics-ML models (3.0):} To overcome the limitations of aforementioned single-paradigm approaches, process-guided or knowledge-guided ML integrates mechanistic insights into data-driven models~\cite{willard2022integrating,karpatne2024knowledge}. This hybrid paradigm embeds physical laws and domain knowledge into machine learning workflows to improve accuracy, generalization, and consistency with fundamental principles such as conservation laws. For instance,
    in lake modeling, process-based components have been combined with recurrent neural networks, yielding better performance for long-term trend predictions by constraining outputs with ecological principles~\cite{jia2019physics,yu2024process}.
\end{itemize}

Building on these computational paradigms, \textbf{foundation models (4.0)} represent the next leap forward. These models, which have already revolutionized domains such as natural language processing (NLP), are set to transform environmental science by enabling holistic, scalable, and integrated modeling approaches. Foundation models excel in their ability to assimilate data from multiple sources and domains, capturing intricate patterns across diverse systems. Their success is also attributed to the pre-training on extensive and diverse datasets, which enables scalability and adaptability from well-studied systems to data-scarce or unseen environments through effective knowledge transfer~\cite{bommasani2021opportunities}.  

\subsection{Emergence of Foundation Models}

Foundation models are large, pre-trained machine learning systems capable of processing diverse modalities. These models have transformed the AI landscape by influencing the physical world, enabling complex reasoning, and facilitating meaningful human interaction. Their development builds on decades of advancements in ML, particularly the evolution of deep learning in the 2010s. Deep learning, driven by the concept of representation learning, allows systems to automatically extract meaningful features directly from raw data, significantly improving performance on complex tasks~\cite{bengio2013representation}. 
Despite their success, early deep learning models were still task-specific, which limited their broader applicability across multiple domains. The true breakthrough came with the advent of transfer learning, which allowed models trained on one task to be adapted to another. Transfer learning enabled the development of general-purpose models that could leverage pre-trained knowledge. Building on this, foundation models such as BERT and GPT expanded transfer learning by training on massive datasets, allowing them to generalize across tasks with minimal tuning~\cite{kalyan2023survey}. 

A key driver of this transformation has been the rise of self-supervised learning, which enables models to learn representative patterns from vast amounts of unlabeled data with pre-defined pre-training tasks~\cite{zhang2024foundation}. 
This is particularly valuable in environmental science, where labeled data is often scarce or incomplete~\cite{ghosh2022robust,pantazis2021focus,hoffmann2023atmodist}. 
Another critical factor in the success of foundation models is their scalability. Models such as GPT-3, with its 175 billion parameters, demonstrate the impact of scaling both data and model size. This scalability enables emergent capabilities like in-context learning, where the model adapts to new tasks simply by receiving natural language prompts~\cite{zhu2023chatgpt,huang2024enviroexam}. In environmental science, this adaptability offers the potential to develop a unified model capable of handling multiple related tasks simultaneously, such as different water quality and quantity variables~\cite{luo2023free,li2024lite}. Such models could serve as powerful tools, offering a comprehensive perspective on diverse processes in dynamic environmental applications. Lastly, architectural innovations, particularly the Transformer architecture introduced by Vaswani et al.~\cite{vaswani2017attention}, have been instrumental in learning complex data patterns. Transformers are adept at capturing long-range dependencies in data, which is crucial for modeling spatial-temporal processes. Their ability to handle large datasets and integrate multi-modal data (e.g., text, images, sensors) makes them invaluable for unified ecosystem modeling~\cite{xu2023multimodal,han2023survey}.

\section{Application-Centric Objectives} 
\label{sec:objective}

\begin{figure}[t]
	\centering
	\includegraphics[width=0.7\textwidth]{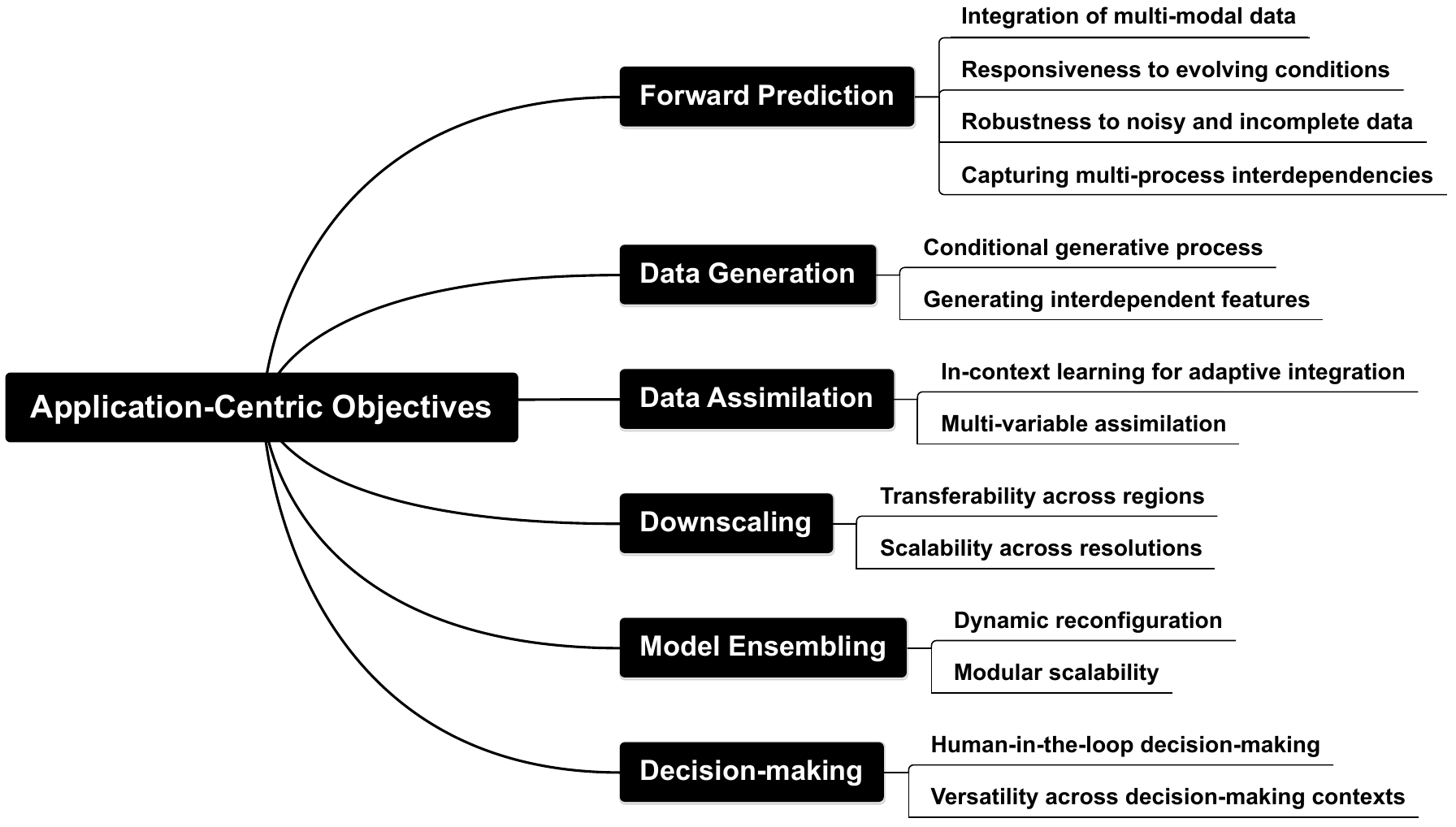} 
	\caption{Application-centric objectives and advancements enabled by foundation models.}
	\label{fig:application_objectives}
\end{figure}

This section examines the application of foundation models in environmental science, focusing on key objectives such as forward prediction, data generation, data assimilation, downscaling, model ensembling, and decision-making. Figure~\ref{fig:application_objectives} highlights their advancements compared to traditional methods.

\subsection{Forward Prediction} 


Forward prediction is a fundamental task in environmental science, focusing on modeling the relationships between input factors, such as climatic drivers or anthropogenic influences, and system characteristics to predict key environmental variables. This approach supports simulations of dynamic environmental processes over time, aiding in tasks like weather forecasting, greenhouse gas emissions modeling, water quality prediction, and tracking the spread of pests.

A specialized subset of forward prediction is forecasting, which emphasizes using historical and current data to predict future states of environmental variables. Forecasting often involves projecting temporal trends, such as long-term changes in climate patterns, air and water quality, or ecosystem health. 
Anomaly detection is another key aspect of forward prediction, focusing on identifying data points or patterns that significantly deviate from expected norms. In environmental science, it plays a crucial role in detecting early signs of critical or unusual changes, enabling timely interventions to prevent or mitigate potential negative impacts on ecosystems and human well-being.

Traditionally, process-based models based on physical principles have been key to environmental applications like climate, weather, and hydrology modeling. Parameterization is often used to handle complex or incompletely understood processes by approximating missing physics~\cite{bennett2021deep}. Parameter calibration approaches and reduced-order models have been developed to enhance model efficiency and reliability while reducing computational costs~\cite{quarteroni2014reduced}.
Data-driven models have emerged as effective alternatives where process-based models are computationally expensive or biased. Deep learning methods excel in extracting complex patterns, with techniques like variational autoencoders (VAEs) used for dimensionality reduction and feature extraction from large datasets~\cite{wu2024novel}. 
For anomaly detection, methods such as isolation forests and clustering models effectively identify outliers in environmental datasets, such as pollutant levels.  
These methods can be complemented by traditional statistical techniques, like
threshold-based methods, to ensure physical plausibility~\cite{blazquez2021review}.

However, the success of these approaches is often shown on clean benchmark datasets for isolated applications. 
Without explicitly leveraging data from multiple sources, they remain limited in 
modeling real-world ecosystems as they often  
involve multiple processes that evolve and interact at different scales. 
In contrast, foundation models present significant opportunities to address these challenges: 
\textbf{(1) Integration of multi-modal data:} Foundation models excel in synthesizing diverse inputs, such as satellite imagery, sensor data, and historical records. For example, ClimaX by Microsoft~\cite{nguyen2023climax} integrates visual, numerical, and textual data to predict climate variables across scales. Huawei’s Pangu-Weather~\cite{bi2023accurate} provides accurate weather forecasts by processing global datasets, demonstrating the capability to model atmospheric dynamics holistically.
\textbf{(2) Responsiveness to evolving conditions:} Foundation models are highly adaptable to changes in spatial and temporal conditions. IBM’s AI climate tools~\cite{mukkavilli2023ai} use fine-tuning on limited regional datasets to model urban heat islands and shifts in precipitation patterns, showcasing rapid adaptability without extensive retraining.
\textbf{(3)~Robustness to noisy and incomplete data:} Environmental datasets often suffer from noise and missing values. Foundation models handle such issues using advanced imputation techniques and architectures capable of learning from incomplete data.  For instance, the large sensor model employs self-supervised learning to effectively impute missing values in extensive sensor datasets~\cite{narayanswamy2024scaling}.  
\textbf{(4) Capturing multi-process interdependencies:} Environmental systems involve complex interdependencies among variables. Models like Prithvi~\cite{Prithvi-100M-preprint} capture interactions between vegetation, soil, and climate. The physics-guided foundation model~\cite{yu2025physics} predicts water temperature and dissolved oxygen in lakes, illustrating how temperature shifts affect oxygen solubility.

\subsection{Data Generation}

Significant data gaps often hinder environmental science applications. For example, process-based models depend on remote sensing data, which can be obscured by noise (e.g., clouds in satellite imagery). Similarly, field-based data, such as soil properties or land use, may be unavailable for certain regions due to the high labor and resource costs required for collection. 

ML-based generative methods aim to address these gaps by generating realistic synthetic data that mimics real-world conditions. Generative models, such as generative adversarial networks (GANs), VAEs, and diffusion models can generate data simulations that closely resemble true observations for many environmental applications, such as forest and crop monitoring
~\cite{martinez2022comparison,nie2022prediction}.  

Foundation models, trained on vast and diverse datasets, present significant opportunities to enhance data generation capabilities in environmental science. These models excel in two key areas: \textbf{(1) Conditional generative process:} Foundation models generate synthetic data tailored to specific conditions by leveraging their ability to encode relationships across diverse features. This is particularly valuable for scenarios with limited observational data or where simulating data under targeted environmental policies is critical. For instance, Deng et al.~\cite{deng2024k2} employed LLaMA to generate domain-specific datasets for data-scarce regions, enabling downstream applications in geoscience. By modifying prompts, LLMs can explore various environmental scenarios, facilitating the simulation of data under different management strategies.
\textbf{(2) Generating interdependent features:} By incorporating multiple data sources, foundation models can effectively capture relationships among diverse variables and processes in the generative process. 
For example, Khanna et al.~\cite{khanna2023diffusionsat} proposed DiffusionSat, which captures internal relationships across multiple data sources, such as high-resolution remote sensing data and text-based inputs, improving performance in generative tasks like temporal generation and inpainting. Similarly, Li et al.~\cite{li2024segment} adapted the Segment Anything model for field mapping in agriculture, demonstrating its ability to integrate diverse environmental data.
However, foundation models are prone to ``hallucination'', where plausible but incorrect data points are generated, especially when they encounter novel data combinations not seen during training. This can lead to misleading conclusions about environmental conditions. 

\subsection{Data Assimilation}

Data assimilation integrates new observational data into physics-based models to improve the accuracy of simulations and predictions. Traditional methods like the Kalman filter have been fundamental for data assimilation in environmental science. For instance, the ensemble Kalman filter (EnKF), widely used for non-linear systems, refines initial conditions by incorporating diverse sources like satellite imagery and ground-based observations, leading to more accurate forecasts in water property prediction~\cite{park2020variable} and crop yield forecasting~\cite{nguyen2019spatial}.

Foundation models have the potential to revolutionize data assimilation in environmental science by integrating diverse observational datasets, including satellite imagery, sensor data, physical simulations, and ground-based measurements. Their versatility and advanced architectures enable improved performance in various applications, such as weather forecasting~\cite{mukkavilli2023ai,nguyen2023climax} and streamflow prediction~\cite{li2024lite}. Key contributions of foundation models to data assimilation include:  
\textbf{(1) In-context learning for adaptive integration:} Foundation models can dynamically adapt to new observations through in-context learning, incorporating recent examples without retraining. This capability allows them to effectively respond to evolving environmental conditions and varying data availability, enhancing accuracy in dynamic scenarios such as extreme weather events or seasonal changes.
\textbf{(2) Multi-variable assimilation:} Unlike traditional methods that typically focus on a single variable, foundation models excel at assimilating diverse environmental variables. For instance, in aquatic systems, they can integrate temperature observations at one time and dissolved oxygen measurements at another~\cite{yu2025physics}, or both field measurements and remote sensing estimations at different frequencies, by leveraging their multi-modal capabilities to synthesize and refine predictions. Models like ClimaX~\cite{nguyen2023climax} illustrate this strength by combining satellite temperature data, weather station observations, and other inputs to produce robust and comprehensive climate predictions.

\subsection{Downscaling}

Downscaling refines large-scale environmental predictions to provide detailed local forecasts by transforming coarse-resolution data from global or regional models into high-resolution outputs. This approach is widely used to bridge the gap between broad-scale models and the specific needs of local-scale environmental management, such as the finer-scale prediction of local temperature and precipitation patterns.  
Traditional methods, including statistical techniques and dynamic modeling, have been extensively used. ML approaches like super-resolution further enhance this process by leveraging large datasets to capture complex patterns often missed by traditional methods~\cite{wang2021deep}. 
However, ML-based methods often struggle to generalize across different spatial regions and scales. 

Foundation models enhance downscaling by utilizing fine-tuning and prompt-based adjustments, allowing outputs to be tailored for specific regions without extensive retraining. The opportunities offered by foundation models include:  
\textbf{(1)~Transferability across regions:} These models can seamlessly adapt knowledge from one region to another, allowing their application across diverse geographic and climatic contexts with minimal adjustments. For example, Dong et al.~\cite{dong2024generative} proposed SMLFR, which incorporates sparse modeling and low-frequency information with learned general patterns to enable satellite image generation across regions.
\textbf{(2) Scalability across resolutions:} These models effectively handle data across various spatial and temporal scales, providing both fine-grained local insights and broader regional analyses, demonstrating their versatility for diverse downscaling tasks. For example, Li et al.~\cite{li2024foundation} explored the potential of GPT-4V in geo-localization, land cover classification, visual question answering, and basic image understanding by leveraging multi-modal and multi-scale data. These strengths of foundation models have also been demonstrated in other applications, such as agriculture~\cite{tan2023promises} and remote sensing~\cite{hong2024spectralgpt,sun2022ringmo,chen2024rsmamba}.

\subsection{Model Ensembling}

Model ensembling in environmental science integrates outputs from multiple predictive models to improve the accuracy and robustness of predictions~\cite{zounemat2021ensemble,shahhosseini2020forecasting}. This technique leverages the strengths of diverse models through approaches such as model blending, which combines various predictions, and post-processing, which refines outputs using observational data. Ensembling is widely applied in environmental disciplines, including weather forecasting, where it aggregates predictions from multiple numerical weather models to enhance forecast precision~\cite{gronquist2021deep,hewage2021deep}. In hydrology, it synthesizes streamflow predictions from different hydrological models, ensuring more reliable and precise forecasts~\cite{troin2021generating,islam2021flood}. Similarly, in climate science, post-processing aligns model outputs with observed climate data to mitigate biases and improve long-term projections~\cite{ahmed2020multi,kochkov2024neural}.

Foundation models capture complex interrelationships among diverse data types and provide greater flexibility in model integration. These models adeptly synthesize datasets such as atmospheric conditions, oceanographic measurements, and terrestrial observations, leveraging advanced algorithms to uncover and utilize relationships across these sources~\cite{rezayi2022agribert,lin2023geogalactica}. Foundation models excel in two key areas: 
\textbf{(1)~Dynamic reconfiguration:} These models enable rapid adjustments to ensemble compositions by incorporating new data or refining underperforming components without extensive retraining. Li et al.~\cite{li2024segment} demonstrate the improvement of SAM's predictive performance in multiple agriculture and climate tasks by incorporating additional natural data.
This adaptability ensures ensembles remain responsive to changing conditions and specific predictive challenges. \textbf{(2)~Modular scalability:} The modular design of foundation models facilitates seamless integration with other predictive models and allows for easy adjustments. This scalability supports diverse applications, ranging from regional hydrological forecasts to global climate modeling, enhancing their effectiveness across various environmental domains. For instance, Kochkov et al.~\cite{kochkov2024neural} proposed NeuralGCM, which integrates a differentiable solver for atmospheric dynamics with machine-learning components, demonstrating competitive performance in weather and climate prediction tasks.

\subsection{Decision-making}

Decision-making in environmental science applies model-driven insights to develop strategies for natural resource management and environmental conservation, focusing on sustainability goals. This process often involves scenario analysis, where models simulate various conditions to evaluate the potential impacts of management strategies, aiding in risk and benefit assessment.

Foundation models streamline decision-making by integrating complex environmental data into actionable strategies. Unlike traditional ML models, which often require intricate reward designs, foundation models leverage extensive pre-training across diverse datasets, enabling them to adapt effectively to varied scenarios without relying on specific reward structures. 
Key advantages of foundation models include:  
\textbf{(1) Human-in-the-loop decision-making:} Foundation models enable meaningful interaction with stakeholders by producing human-interpretable outputs, such as visualizations that clarify predictions and recommendations. This capability facilitates transparent and collaborative decision-making processes, making the insights accessible to policymakers and community stakeholders~\cite{tan2023promises}.  
\textbf{(2) Versatility across decision-making contexts:} Foundation models exhibit strong adaptability, handling diverse tasks such as linking global climate data to localized weather predictions or combining new environmental monitoring data with established models. For example, models like ClimaX~\cite{nguyen2023climax}  support a wide range of decision-making applications, from biodiversity conservation to water resource planning in dynamic conditions.

\begin{figure}[t]
	\centering
	\includegraphics[width=\textwidth]{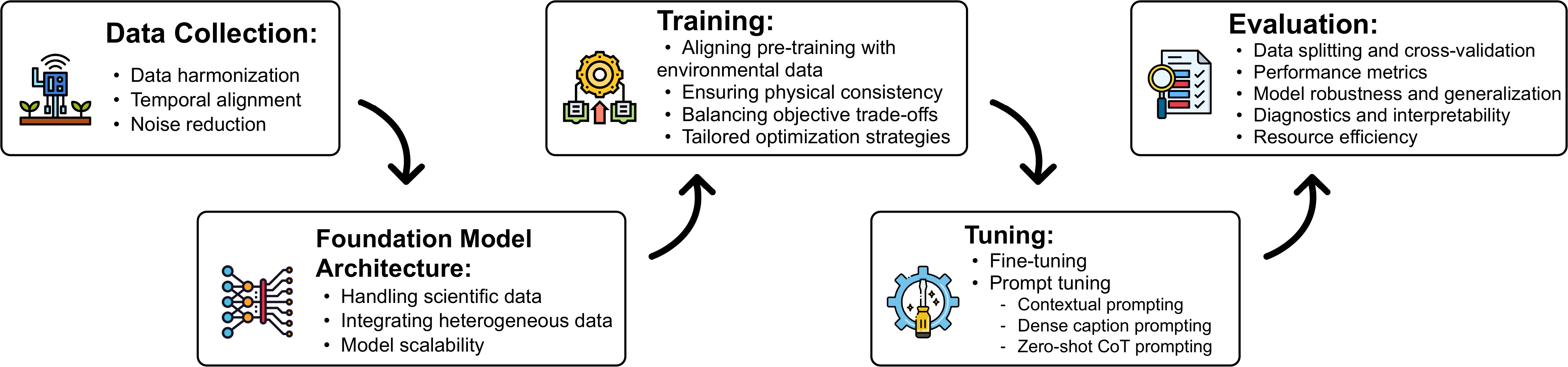} 
	\caption{Model design workflow for foundation models in environmental science.}
	\label{fig:method_design}
\end{figure}

\section{Method Design}
\label{sec:method}

This section outlines the methodologies for designing foundation models in environmental science, as summarized in Figure~\ref{fig:method_design}. It systematically examines key stages of the development process, including data collection, architecture design, training, tuning, and evaluation. 


\subsection{Data Collection}

Building robust foundation models for environmental science requires addressing challenges in data collection and processing to ensure diverse, large-scale, and high-quality datasets. Capturing a broad range of environmental conditions supports generalization across varied contexts, necessitating datasets from ecosystems and climates such as tropical rainforests, arid deserts, and polar regions. Simulated data can effectively fill gaps where observational data is sparse, enabling models to incorporate rare or underrepresented phenomena.  

Integrating heterogeneous datasets involves resolving variations in format, resolution, and temporal frequency. For example, satellite imagery provides high-resolution spatial data but often lacks temporal continuity, whereas ground sensors offer detailed time-series data with limited spatial coverage. Missing or noisy data, such as cloud-obscured satellite images or sensor malfunctions, further complicate integration~\cite{dong2024generative,nguyen2023climax}. Key preprocessing steps include:  
\textbf{\textit{Data harmonization:}} Standardizing units and aligning spatial resolutions to ensure consistency across datasets.  
\textbf{\textit{Temporal alignment:}} Synchronizing datasets with different sampling frequencies to maintain temporal consistency.
\textbf{\textit{Noise reduction:}} Techniques such as wavelet transforms and median filtering 
remove distortions introduced during data collection.  

Identifying relevant variables ensures an accurate representation of environmental processes. Variables such as temperature, precipitation, vegetation indices, and pollutant concentrations are commonly prioritized. The data collected are often unbalanced in diverse regions. To address spatial biases and overrepresented regions, techniques like spatial weighting are often adopted to ensure balanced and representative model performance.

Access to high-quality data sources enhances building foundation models:   
\textbf{Meteorological data:} The National Oceanic and Atmospheric Administration (NOAA) and the European Centre for Medium-Range Weather Forecasts (ECMWF) provide comprehensive weather and climate datasets for short-term forecasting and long-term climate modeling.  
\textbf{Geological and hydrological data:} The United States Geological Survey (USGS) offers datasets on geological conditions, natural hazards, and water quality.  
\textbf{Remote sensing data:} Resources like the National Land Cover Database (NLCD) and HydroLAKES provide high-resolution spatial data on land cover and water body distributions\footnote{URLs for the mentioned resources: NOAA (\url{https://www.noaa.gov}), ECMWF (\url{https://www.ecmwf.int}), USGS (\url{https://www.usgs.gov}), NLCD (\url{https://www.mrlc.gov}), HydroLAKES (\url{https://www.hydrosheds.org/pages/hydrolakes}).}.

\subsection{Foundation Model Architecture}

Designing foundation model architectures for environmental science involves addressing challenges related to scientific data complexity, data heterogeneity, and scalability across regions and timescales.

\noindent \textit{\textbf{Handling scientific data:}}
Environmental data is often complex and domain-specific, requiring specialized model architectures. Pre-trained models, such as encoder-decoder transformers, are effective for text-based datasets. Fine-tuning these models on environmental data enhances their ability to analyze reports, extract insights, and support decision-making~\cite{deng2024k2}. For instance, T5 models fine-tuned on climate data can generate hypotheses or support adaptation strategies~\cite{luo2023free,lacoste2024geo}.
For time-series data, temporal transformers capture temporal patterns and incorporate contextual factors, making them suitable for forecasting seasonal trends or weather patterns~\cite{nguyen2023climax}. Physics-guided foundation model integrates physical principles, such as energy or water cycle laws, to ensure scientifically consistent predictions, particularly in scenarios with limited data~\cite{karpatne2024knowledge}.

\noindent \textit{\textbf{Integrating heterogeneous data:}}
Environmental datasets often differ in format, resolution, and frequency, requiring models that integrate diverse data sources. Multi-modal transformers combine inputs such as satellite imagery, sensor readings, and numerical data into unified representations for analysis. For example, these models can integrate satellite and ground-based data to assess land cover or air quality changes~\cite{lin2023geogalactica,li2024segment}.
Spatial-temporal graph neural networks connect spatial and temporal data, such as linking weather station time-series data with geographic information to model regional dynamics~\cite{luo2023free,karpatne2024knowledge}. Adaptive transformers streamline preprocessing by aligning datasets with different resolutions and frequencies, simplifying data integration~\cite{wang2023siamhrnet}.

\noindent \textit{\textbf{Model scalability:}}
Scalability is essential for addressing multi-region and multi-scale challenges in environmental science. Fourier neural operators (FNO)~\cite{li2020fourier} efficiently model cross-scale interactions, making them ideal for tasks like predicting regional rainfall while considering global climate patterns. For high-resolution data, vision Transformers and hierarchical models like Swin Transformers analyze satellite imagery to monitor deforestation or urbanization~\cite{tan2023promises,wang2023siamhrnet}. Transfer learning supports scalability by adapting models trained on global datasets to specific regions or contexts. For example, a climate model trained globally can be fine-tuned for regional biodiversity monitoring or flood prediction~\cite{tan2023promises,wang2024gpt}.

\subsection{Training}

Training or pre-training foundation models for environmental science requires strategies to capture the complexities of environmental systems and facilitate generalization across diverse tasks and datasets.

\noindent \textit{\textbf{Aligning pre-training with environmental data:} } 
Pre-training tasks should reflect the unique characteristics of environmental science. Instead of relying on generic objectives, tasks can focus on reconstructing physical variables (e.g., temperature, precipitation) or modeling multi-scale environmental interactions~\cite{yu2025physics}. Self-supervised learning methods, such as masked autoencoders~\cite{he2022masked}, allow models to leverage unlabeled datasets by predicting missing components. Contrastive learning, which distinguishes between similar and dissimilar contexts, enhances generalization to downstream tasks. These methods ensure that models learn representations aligned with environmental dynamics, including climate variability and ecosystem processes~\cite{zhu2024foundations,li2024foundation,yu2025physics}.

\noindent \textit{\textbf{Ensuring physical consistency:}  }
Incorporating physical laws during training improves the reliability of predictions. Physical loss functions, such as enforcing energy or mass conservation, help maintain adherence to scientific principles~\cite{karpatne2024knowledge}. For instance, in hydrology, these loss functions can ensure that predictions respect water balance equations~\cite{yu2025physics}. Hybrid models, which combine machine learning with physics-based simulations, leverage the strengths of both approaches for enhanced consistency. Domain adaptation techniques transfer knowledge from data-rich to data-scarce regions~\cite{willard2022integrating,karpatne2024knowledge}.

\noindent \textit{\textbf{Balancing objective trade-offs:}  }
Training should balance predictive accuracy with adherence to physical principles. Multi-objective optimization techniques, such as Pareto optimization, optimize multiple criteria simultaneously, ensuring a balance between error minimization and physical consistency. Regularization methods, like physics-informed priors, help maintain this balance without overfitting. Advanced optimizers, such as AdamW, stabilize training by effectively handling large parameter spaces and preventing weight decay issues.

\noindent \textit{\textbf{Tailored optimization strategies:} }
Training strategies tailored to environmental data improve performance:
\textbf{Curriculum learning:} Start with simpler tasks, such as single-variable predictions, and progressively incorporate complex tasks, such as multi-variable interactions. This helps models build foundational knowledge before addressing advanced scenarios.
 \textbf{Multi-scale training:} Environmental processes operate at varying spatial and temporal scales. Training on datasets spanning local, regional, and global scales ensures generalization across different resolutions~\cite{li2020fourier,nguyen2023climax}.
\textbf{Gradient clipping:} Stabilize training and prevent exploding gradients in large-scale models by limiting the magnitude of updates.

\subsection{Tuning}

Tuning adapts pre-trained foundation models to specific tasks and datasets in environmental science, aligning them with domain-specific challenges for practical applications. Two main approaches are fine-tuning and prompt tuning, each targeting distinct aspects of model customization.

\noindent \textit{\textbf{Fine-tuning:}}
Fine-tuning involves retraining pre-trained models on task-specific datasets to improve their performance for specialized applications, such as climate modeling, air quality analysis, or biodiversity studies. A common strategy is layer-wise adaptation, where earlier layers—responsible for general features—are minimally updated, while later task-specific layers are fine-tuned more aggressively. Freezing base layers during the initial training phases reduces computational costs and prevents overfitting, especially when working with smaller task-specific datasets. Fine-tuning also requires careful hyperparameter optimization to balance generalization and accuracy~\cite{zhu2024foundations,li2024foundation}. Beyond standard adaptations, domain knowledge can be integrated to ensure physically valid predictions~\cite{yu2025physics}. Lastly, models are often fine-tuned on localized datasets to enhance predictions for specific regions. For instance, a model pre-trained on global climate data can be fine-tuned with regional meteorological datasets to optimize predictions for localized phenomena, such as drought forecasting or flood risk assessment~\cite{islam2021flood}. 

\noindent \textit{\textbf{Prompt tuning:}}
Prompt tuning adapts pre-trained models by structuring or modifying input prompts to guide their outputs without altering model parameters. This approach is especially valuable for domain-specific challenges, such as land use classification, climate forecasting, or species identification, where minimal retraining is preferred.
\textbf{Contextual prompting:} Contextual prompting adds task-specific context to the input to guide the model’s understanding. For example, a prompt for species classification might include, “Each species has a unique scientific name composed of genus and species,” enhancing the model’s interpretive accuracy while preserving its original structure.
\textbf{Dense caption prompting:} This technique generates detailed descriptions of input features, such as satellite imagery, and incorporates them into the task prompt. For example, in land cover analysis, the model first describes features like vegetation and water bodies before addressing questions on environmental changes. This approach enriches data understanding for complex tasks.
\textbf{Zero-shot chain of thought (CoT) prompting:} CoT prompting encourages step-by-step reasoning, which is critical for multi-factor analyses. For example, in evaluating deforestation impacts, a prompt might guide the model to consider sequential effects on carbon storage, local temperatures, and biodiversity. 

\subsection{Evaluation}

Effective evaluation of foundation models in environmental science focuses on their ability to address unique challenges, such as multi-modal data integration, scalability, and generalization across diverse environmental tasks and datasets.

\noindent \textit{\textbf{Data splitting and cross-validation:}}
Time-based and region-based splits ensure models account for temporal dynamics and spatial heterogeneity without introducing data leakage. K-fold cross-validation assesses transferability across varied datasets, while domain-specific methods, such as leave-one-region-out validation, evaluate performance on unobserved geographic areas~\cite{karpatne2024knowledge}.

\noindent \textit{\textbf{Performance metrics:}} 
Beyond conventional regression metrics like root mean square error, multi-modal alignment scores, such as contrastive loss in joint embeddings, assess the model’s ability to integrate text, imagery, and numerical data. For downstream environmental applications, extrinsic evaluation focuses on metrics such as F1 scores for species classification or intersection-over-union (IoU) for land-cover segmentation, ensuring task-specific adaptation~\cite{zhu2024foundations,li2024foundation}.

\noindent \textit{\textbf{Model robustness and generalization:}} Foundation models should demonstrate robustness across diverse environmental conditions and unseen data domains. 
Transferability tests evaluate performance when the model is fine-tuned or adapted to new regions or phenomena, such as applying a global weather model to predict localized flood patterns~\cite{islam2021flood}. Out-of-distribution testing assesses generalization to novel modalities or sparsely observed datasets, a critical capability for applications like rare species identification or extreme event prediction~\cite{zhu2024foundations}.

\noindent \textit{\textbf{Diagnostics and interpretability:}} Evaluating foundation models involves diagnosing biases and interpreting decision-making processes. Feature importance tools help interpret the contribution of variables across multi-modal inputs~\cite{mangalathu2020failure,yu2025physics}. They can reveal whether vegetation indices or precipitation data dominate predictions in drought forecasting tasks. Residual analysis uncovers systematic biases, such as overgeneralizing patterns across diverse geographic areas~\cite{bi2022pangu}.

\noindent \textit{\textbf{Resource efficiency:}}
Foundation models are expected to efficiently process large-scale environmental datasets, such as high-resolution satellite imagery or climate simulations. Evaluation should include scalability metrics, such as runtime per dataset size or memory usage for multi-modal tasks. Efficient adaptation techniques, such as parameter-efficient tuning (e.g., LoRA or prompt tuning), are evaluated based on their ability to minimize computational overhead while maintaining performance on environmental applications~\cite{bommasani2021opportunities,li2024foundation}.

\begin{table}[!t]
\centering
\caption{Application-centric objectives and methods for foundation models.}
\begin{adjustbox}{max width=\textwidth}
\renewcommand{\arraystretch}{1.5} 
\begin{tabular}{|>{\centering\arraybackslash}m{3.5cm}|>{\centering\arraybackslash}m{3.2cm}|>{\centering\arraybackslash}m{4cm}|>{\centering\arraybackslash}m{3cm}|>{\centering\arraybackslash}m{3cm}|>{\centering\arraybackslash}m{3.2cm}|}
\hline
\textbf{3 Objectives} & \textbf{4.1 Data Collection} & \textbf{4.2 Foundation Model Architecture} & \textbf{4.3 Training} & \textbf{4.4 Tuning} & \textbf{4.5 Evaluation} \\ \hline
3.1 Forward Prediction & \cite{nguyen2023climax,li2024segment,bi2022pangu,kochkov2024neural,bodnar2024aurora,guo2024skysense,dong2024generative,ren2024watergpt,stevens2024bioclip} & \cite{li2024foundation,hong2024spectralgpt,chen2024rsmamba,nguyen2023climax,rezayi2022agribert,lin2023geogalactica,bi2022pangu,lacoste2024geo,bodnar2024aurora,guo2024skysense,dong2024generative,ren2024watergpt,stevens2024bioclip} & \cite{li2024foundation,deng2024k2,nguyen2023climax,bi2022pangu,kochkov2024neural,lacoste2024geo,guo2024skysense,dong2024generative,ren2024watergpt,stevens2024bioclip,yu2025physics} & \cite{li2024foundation,sun2022ringmo,nguyen2023climax,luo2023free,bi2022pangu,kochkov2024neural,lacoste2024geo,guo2024skysense,dong2024generative,ren2024watergpt,stevens2024bioclip,yu2025physics} & \cite{nguyen2023climax,bi2022pangu,kochkov2024neural,lacoste2024geo,stevens2024bioclip,yu2025physics} \\ \hline
3.2 Data Generation & \cite{nguyen2023climax,li2024segment,bi2022pangu,kochkov2024neural,guo2024skysense,dong2024generative,stevens2024bioclip,li2024using} & \cite{deng2024k2,khanna2023diffusionsat,kochkov2024neural,guo2024skysense,dong2024generative,stevens2024bioclip,li2024using} & \cite{khanna2023diffusionsat,kochkov2024neural,cong2022satmae,reed2023scale,dong2024generative,stevens2024bioclip} & \cite{dong2024generative,stevens2024bioclip} & \\ \hline
3.3 Data Assimilation & \cite{nguyen2023climax,li2024segment,bi2022pangu} & \cite{mukkavilli2023ai,nguyen2023climax} & \cite{nguyen2023climax,yu2025physics} & \cite{li2024lite,yu2025physics} & \cite{yu2025physics} \\ \hline
3.4 Downscaling & \cite{nguyen2023climax,li2024segment,bi2022pangu,kochkov2024neural,dong2024generative} & \cite{tan2023promises,hong2024spectralgpt,nguyen2023climax,khanna2023diffusionsat,kochkov2024neural,guo2024skysense,dong2024generative} & \cite{sun2022ringmo,khanna2023diffusionsat,kochkov2024neural,cong2022satmae,reed2023scale,dong2024generative} & \cite{khanna2023diffusionsat,kochkov2024neural,dong2024generative} & \\ \hline
3.5 Model Ensembling & & \cite{rezayi2022agribert,lin2023geogalactica,kochkov2024neural} & & \cite{li2024segment} & \\ \hline
3.6 Decision-making & & \cite{nguyen2023climax,tan2023promises} & & & \\ \hline
\end{tabular}
\end{adjustbox}
\label{tab:Overview}
\end{table}

\section{Discussions}
\label{sec:discussion}

Table~\ref{tab:Overview} summarizes the applications of foundation models in environmental science. Areas with extensive research activity indicate current hotspots, while those with fewer contributions highlight challenges or opportunities for further exploration. This section reflects on the current state of foundation models in this field, their limitations, and potential directions for future research.

Foundation models offer immense potential for environmental science but remain in the early stages of their application. By integrating diverse data sources such as satellite imagery, sensor measurements, and historical records, these models provide a holistic understanding of environmental systems. Pre-training on large-scale datasets enables these models to quickly adapt to new tasks. 
Their adaptability, driven by techniques like prompt engineering and tuning, enables meaningful insights even from incomplete or variable data. Emerging technologies like retrieval-augmented generation (RAG) and in-context learning also show promise for incorporating external knowledge into real-time decision-making.

However, foundation models face notable challenges that need to be addressed to fully unlock their potential. One key obstacle in scientific discovery is the lack of trust in these models due to their ``black-box'' nature. Addressing this challenge requires advancements in the explainability of their predictions. 
Uncertainty quantification for these models is also underdeveloped, yet it is vital for high-stakes applications such as disaster preparedness, climate mitigation, and resource management. Modeling rare extreme events, such as floods and heatwaves, is particularly difficult due to sparse data and the complexity of integrating physical constraints. Computational efficiency poses a significant barrier, especially for high-resolution simulations over large regions and long periods. 

Looking forward, several opportunities exist to advance the development and application of foundation models in environmental science. Embedding domain knowledge can improve physical consistency and enhance model interpretability, addressing concerns around explainability and trust. Open-access datasets and pre-trained models can democratize research and foster interdisciplinary collaboration, broadening participation and accelerating innovation. Active learning offers a practical way to enhance model performance by prioritizing data collection in regions with high uncertainty or limited representation. Integrating decision-making frameworks into foundation models has the potential to transform them from predictive tools into actionable systems, supporting adaptive management strategies for water resources, disaster mitigation, and more. Addressing long-term error accumulation through recalibration, ensemble modeling, and hybrid approaches can further improve the reliability of extended forecasts. Finally, foundation models hold significant potential for scientific discovery by uncovering novel patterns, generating hypotheses, and synthesizing interdisciplinary knowledge, advancing our understanding of complex environmental systems.

\section{Conclusion}

In this survey, we explored the transformative potential of foundation models in environmental science, reviewing their applications and detailing methodologies for their development. These models, with their ability to integrate diverse data, capture complex spatiotemporal dynamics, and adapt flexibly, represent a significant leap forward in addressing pressing environmental challenges. By fostering interdisciplinary collaboration and advancing open-access resources, foundation models can inspire innovative solutions, enhance scientific discovery, and support informed decision-making for sustainable development. We anticipate that this work will serve as a foundation for future research, driving progress toward a more resilient and informed global community.

%
%
%
\clearpage

\begin{thebibliography}{10}
\providecommand{\url}[1]{\texttt{#1}}
\providecommand{\urlprefix}{URL }
\providecommand{\doi}[1]{https://doi.org/#1}

\bibitem{ahmed2020multi}
Ahmed, K., et~al.: Multi-model ensemble predictions of precipitation and
  temperature using machine learning algorithms. Atmospheric Research  (2020)

\bibitem{bengio2013representation}
Bengio, Y., et~al.: Representation learning: A review and new perspectives.
  IEEE transactions on pattern analysis and machine intelligence  (2013)

\bibitem{bennett2021deep}
Bennett, A., et~al.: Deep learned process parameterizations provide better
  representations of turbulent heat fluxes in hydrologic models. WRR  (2021)

\bibitem{bergen2019machine}
Bergen, K.J., et~al.: Machine learning for data-driven discovery in solid earth
  geoscience. Science  (2019)

\bibitem{bi2022pangu}
Bi, K., et~al.: Pangu-weather: A 3d high-resolution model for fast and accurate
  global weather forecast. arXiv  (2022)

\bibitem{bi2023accurate}
Bi, K., et~al.: Accurate medium-range global weather forecasting with 3d neural
  networks. Nature  (2023)

\bibitem{blazquez2021review}
Bl{\'a}zquez-Garc{\'\i}a, A., et~al.: A review on outlier/anomaly detection in
  time series data. ACM computing surveys (CSUR)  (2021)

\bibitem{bodnar2024aurora}
Bodnar, C., et~al.: Aurora: A foundation model of the atmosphere. arXiv  (2024)

\bibitem{bommasani2021opportunities}
Bommasani, R., et~al.: On the opportunities and risks of foundation models.
  arXiv preprint arXiv:2108.07258  (2021)

\bibitem{chen2024rsmamba}
Chen, K., et~al.: Rsmamba: Remote sensing image classification with state space
  model. IEEE Geoscience and Remote Sensing Letters  (2024)

\bibitem{cong2022satmae}
Cong, Y., et~al.: Satmae: Pre-training transformers for temporal and
  multi-spectral satellite imagery. NeurIPS  (2022)

\bibitem{deng2024k2}
Deng, C., et~al.: K2: A foundation language model for geoscience knowledge
  understanding and utilization. In: WSDM (2024)

\bibitem{dong2024generative}
Dong, Z., et~al.: Generative convnet foundation model with sparse modeling and
  low-frequency reconstruction for remote sensing image interpretation. IEEE
  Transactions on Geoscience and Remote Sensing  (2024)

\bibitem{ghosh2022robust}
Ghosh, R., et~al.: Robust inverse framework using knowledge-guided
  self-supervised learning: An application to hydrology. In: KDD (2022)

\bibitem{gronquist2021deep}
Gr{\"o}nquist, P., ohters: Deep learning for post-processing ensemble weather
  forecasts. Philosophical Transactions of the Royal Society A  (2021)

\bibitem{guo2024skysense}
Guo, X., et~al.: Skysense: A multi-modal remote sensing foundation model
  towards universal interpretation for earth observation imagery. In: CVPR
  (2024)

\bibitem{han2023survey}
Han, X., et~al.: A survey of transformer-based multimodal pre-trained modals.
  Neurocomputing  (2023)

\bibitem{he2022masked}
He, K., et~al.: Masked autoencoders are scalable vision learners. In: CVPR
  (2022)

\bibitem{hewage2021deep}
Hewage, P., ohters: Deep learning-based effective fine-grained weather
  forecasting model. Pattern Analysis and Applications  (2021)

\bibitem{hipsey2019general}
Hipsey, M.R., et~al.: A general lake model (glm 3.0) for linking with
  high-frequency sensor data from the global lake ecological observatory
  network (gleon). Geoscientific Model Development  (2019)

\bibitem{hoffmann2023atmodist}
Hoffmann, S., Lessig, C.: Atmodist: Self-supervised representation learning for
  atmospheric dynamics. Environmental Data Science  (2023)

\bibitem{hong2024spectralgpt}
Hong, D., et~al.: Spectralgpt: Spectral remote sensing foundation model. IEEE
  Transactions on Pattern Analysis and Machine Intelligence  (2024)

\bibitem{hospedales2021meta}
Hospedales, T., et~al.: Meta-learning in neural networks: A survey. IEEE
  transactions on pattern analysis and machine intelligence  (2021)

\bibitem{huang2024enviroexam}
Huang, Y., et~al.: Enviroexam: Benchmarking environmental science knowledge of
  large language models. arXiv preprint arXiv:2405.11265  (2024)

\bibitem{islam2021flood}
Islam, A.R.M.T., et~al.: Flood susceptibility modelling using advanced ensemble
  machine learning models. Geoscience Frontiers  (2021)

\bibitem{Prithvi-100M-preprint}
Jakubik, J., et~al.: {Foundation Models for Generalist Geospatial Artificial
  Intelligence}. Preprint Available on arxiv:2310.18660  (2023)

\bibitem{jia2019physics}
Jia, X., et~al.: Physics guided rnns for modeling dynamical systems: A case
  study in simulating lake temperature profiles. In: SDM (2019)

\bibitem{kalyan2023survey}
Kalyan, K.S.: A survey of gpt-3 family large language models including chatgpt
  and gpt-4. Natural Language Processing Journal  (2023)

\bibitem{karpatne2024knowledge}
Karpatne, A., Jia, X., Kumar, V.: Knowledge-guided machine learning: Current
  trends and future prospects. arXiv preprint arXiv:2403.15989  (2024)

\bibitem{karpatne2018machine}
Karpatne, A., et~al.: Machine learning for the geosciences: Challenges and
  opportunities. IEEE Trans. on Knowledge and Data Engineering  (2018)

\bibitem{khanna2023diffusionsat}
Khanna, S., et~al.: Diffusionsat: A generative foundation model for satellite
  imagery. arXiv preprint arXiv:2312.03606  (2023)

\bibitem{kochkov2024neural}
Kochkov, D., et~al.: Neural general circulation models for weather and climate.
  Nature  (2024)

\bibitem{lacoste2024geo}
Lacoste, A., et~al.: Geo-bench: Toward foundation models for earth monitoring.
  NeurIPS  (2024)

\bibitem{li2024lite}
Li, H., et~al.: Lite: Modeling environmental ecosystems with multimodal large
  language models. arXiv preprint arXiv:2404.01165  (2024)

\bibitem{li2024foundation}
Li, J., et~al.: Foundation models in smart agriculture: Basics, opportunities,
  and challenges. Computers and Electronics in Agriculture  (2024)

\bibitem{li2024using}
Li, N., et~al.: Using llms to build a database of climate extreme impacts. In:
  ACL Workshop ClimateNLP (2024)

\bibitem{li2024segment}
Li, W., et~al.: Segment anything model can not segment anything: Assessing ai
  foundation model’s generalizability in permafrost mapping. Remote Sensing
  (2024)

\bibitem{li2020fourier}
Li, Z., et~al.: Fourier neural operator for parametric partial differential
  equations. arXiv  (2020)

\bibitem{lin2023geogalactica}
Lin, Z., et~al.: Geogalactica: A scientific large language model in geoscience.
  arXiv preprint arXiv:2401.00434  (2023)

\bibitem{luo2023free}
Luo, S., Ni, J., Chen, S., et~al.: Free: The foundational semantic recognition
  for modeling environmental ecosystems. arXiv preprint arXiv:2311.10255
  (2023)

\bibitem{mangalathu2020failure}
Mangalathu, S., et~al.: Failure mode and effects analysis of rc members based
  on machine-learning-based shapley additive explanations (shap) approach.
  Engineering Structures  (2020)

\bibitem{martinez2022comparison}
Martinez, J., et~al.: A comparison of cloud removal methods for deforestation
  monitoring in amazon rainforest. ISPRS  (2022)

\bibitem{moorcroft2001method}
Moorcroft, P.R., et~al.: A method for scaling vegetation dynamics: the
  ecosystem demography model (ed). Ecological monographs  (2001)

\bibitem{mukkavilli2023ai}
Mukkavilli, S.K., et~al.: Ai foundation models for weather and climate:
  Applications, design, and implementation. arXiv preprint arXiv:2309.10808
  (2023)

\bibitem{narayanswamy2024scaling}
Narayanswamy, G., et~al.: Scaling wearable foundation models. arXiv preprint
  arXiv:2410.13638  (2024)

\bibitem{nguyen2019spatial}
Nguyen, L.H., et~al.: Spatial-temporal multi-task learning for within-field
  cotton yield prediction. In: PAKDD. Springer (2019)

\bibitem{nguyen2023climax}
Nguyen, T., et~al.: Climax: A foundation model for weather and climate. ICML
  (2025)

\bibitem{nie2022prediction}
Nie, J., et~al.: Prediction of liquid magnetization series data in agriculture
  based on enhanced cgan. Frontiers in plant science  (2022)

\bibitem{pantazis2021focus}
Pantazis, O., et~al.: Focus on the positives: Self-supervised learning for
  biodiversity monitoring. In: ICCV (2021)

\bibitem{park2020variable}
Park, S., et~al.: Variable update strategy to improve water quality forecast
  accuracy in multivariate data assimilation using the ensemble kalman filter.
  Water research  \textbf{176},  115711 (2020)

\bibitem{quarteroni2014reduced}
Quarteroni, A., Rozza, G., et~al.: Reduced order methods for modeling and
  computational reduction, vol.~9. Springer (2014)

\bibitem{reed2023scale}
Reed, C.J., et~al.: Scale-mae: A scale-aware masked autoencoder for multiscale
  geospatial representation learning. In: CVPR (2023)

\bibitem{ren2024watergpt}
Ren, Y., et~al.: Watergpt: Training a large language model to become a
  hydrology expert. Water  (2024)

\bibitem{rezayi2022agribert}
Rezayi, S., et~al.: Agribert: Knowledge-infused agricultural language models
  for matching food and nutrition. In: IJCAI (2022)

\bibitem{shahhosseini2020forecasting}
Shahhosseini, M., Hu, G., Archontoulis, S.V.: Forecasting corn yield with
  machine learning ensembles. Frontiers in Plant Science  \textbf{11}, ~1120
  (2020)

\bibitem{stevens2024bioclip}
Stevens, S., et~al.: Bioclip: A vision foundation model for the tree of life.
  In: CVPR. pp. 19412--19424 (2024)

\bibitem{sun2022ringmo}
Sun, X., et~al.: Ringmo: A remote sensing foundation model with masked image
  modeling. IEEE Trans. Geosci Remote  (2022)

\bibitem{tan2023promises}
Tan, C., et~al.: On the promises and challenges of multimodal foundation models
  for geographical, environmental, agricultural, and urban planning
  applications. arXiv preprint arXiv:2312.17016  (2023)

\bibitem{troin2021generating}
Troin, M., et~al.: Generating ensemble streamflow forecasts: A review of
  methods and approaches over the past 40 years (2021)

\bibitem{vaswani2017attention}
Vaswani, A., et~al.: Attention is all you need. NeurIPS  \textbf{30} (2017)

\bibitem{wang2021deep}
Wang, F., et~al.: Deep learning for daily precipitation and temperature
  downscaling. Water Resources Research  (2021)

\bibitem{wang2024gpt}
Wang, S., et~al.: Gpt, large language models and generative artificial
  intelligence models in geospatial science: a systematic review. Digital Earth
   (2024)

\bibitem{wang2023siamhrnet}
Wang, Z., et~al.: Siamhrnet-ocr: a novel deforestation detection model with
  high-resolution imagery and deep learning. Remote Sensing  (2023)

\bibitem{willard2022integrating}
Willard, J., et~al.: Integrating scientific knowledge with machine learning for
  engineering and environmental systems. ACM Computing Surveys  (2022)

\bibitem{wu2024novel}
Wu, Q., et~al.: A novel deep learning framework with variational auto-encoder
  for indoor air quality prediction. Frontiers of Environmental Sci \& Eng
  (2024)

\bibitem{xu2023multimodal}
Xu, P., et~al.: Multimodal learning with transformers: A survey. IEEE Trans. on
  Pattern Analysis and Machine Intelligence  \textbf{45}(10),  12113--12132
  (2023)

\bibitem{yu2024process}
Yu, R., et~al.: Adaptive process-guided learning: An application in predicting
  lake do concentrations. In: ICDM (2024)

\bibitem{yu2025physics}
Yu, R., et~al.: Physics-guided foundation model for scientific discovery: An
  application to aquatic science. arXiv preprint arXiv:2502.06084  (2025)

\bibitem{zeyang2022environmental}
Zeyang, W., et~al.: Environmental computing: Concept, evolution, and
  challenges. Journal of Tsinghua University (Science and Technology)
  \textbf{62}(12),  1839--1850 (2022)

\bibitem{zhang2024foundation}
Zhang, M., et~al.: Foundation model for generalist remote sensing intelligence:
  potentials and prospects. Science Bulletin  (2024)

\bibitem{zhu2023chatgpt}
Zhu, J.J., et~al.: Chatgpt and environmental research. Environmental Science \&
  Technology  (2023)

\bibitem{zhu2024foundations}
Zhu, X.X., et~al.: On the foundations of earth and climate foundation models.
  arXiv preprint arXiv:2405.04285  (2024)

\bibitem{zhuang2020comprehensive}
Zhuang, F., et~al.: A comprehensive survey on transfer learning. Proceedings of
  the IEEE  (2020)

\bibitem{zounemat2021ensemble}
Zounemat-Kermani, M., et~al.: Ensemble machine learning paradigms in hydrology:
  A review. Journal of Hydrology  (2021)

\end{thebibliography}

\end{document}